%% file: wacv_camera_ready.tex
\documentclass[10pt,twocolumn,letterpaper]{article}

\usepackage[pagenumbers]{wacv} % To force page numbers, e.g. for an arXiv version

\usepackage{xcolor}
\usepackage{bbm}
\newcommand{\rev}[1]{#1} 

\definecolor{wacvblue}{rgb}{0.21,0.49,0.74}
\usepackage[pagebackref,breaklinks,colorlinks,allcolors=wacvblue]{hyperref}
\usepackage{booktabs}
\usepackage{tabularx}
\usepackage{multirow}

\def\wacvPaperID{2758} % *** Enter the WACV Paper ID here
\def\confName{WACV}
\def\confYear{2027}

\title{RigidGS: Geometry-Aware Object Decoupling for Rigid-Body Interaction in 3D Gaussian Splatting}

\author{
Paul Sansah Gyreyiri \! and \! Diksha Shukla,  Senior Member, IEEE\\
University of Wyoming, Laramie, WY, USA\\
{\tt\small pgyreyir@uwyo.edu, dshukla@uwyo.edu}
}

\begin{document}
\maketitle
\begin{abstract}
Interactive VR/MR applications demand photorealistic 3D representations that support real-time rendering and object-level manipulation. While 3D Gaussian Splatting (3DGS) achieves high rendering quality, it represents scenes as unstructured, monolithic primitives. Existing object-aware 3DGS methods jointly optimize geometry and semantic states but suffer from two key failure modes: noisy foundation-model masks on complex geometry, and \emph{label--geometry coupling}, in which co-optimized identity gradients mislabel boundary Gaussians. We introduce RigidGS, an end-to-end pipeline that decouples geometric learning from object identity. We propose a depth- and temporally-aware mask refinement module that addresses two complementary SAM2 failures: over-inclusion on thin structures and under-inclusion on multimodal-depth objects. We then determine object identities entirely upstream via COLMAP projection voting and freeze them during training, directing mask supervision solely toward geometric optimization. Refined masks alone substantially improve competing methods, yet a residual gap in label-geometry coupling persists that our fixed-identity design eliminates. RigidGS outperforms prior object-aware 3DGS methods on LLFF/NVOS, Mip-NeRF~360, and iNeRF360, improving segmentation and boundary IoU by 7.7 and 14.5 points on LLFF/NVOS. Finally, RigidGS enables real-time rigid-body manipulation on Meta Quest~2 without mesh extraction. Our code and project page are available at \url{https://gyreyirisansah.github.io/rigidgs}.
\end{abstract}
%%%%%%%%%%%%%%%%%%%%%%%%%%%%%%%%%%%%%%%%%%%%%%%%%%%%%%%%%%%%%%%%%%%%%%%%%%%%%%%%%%%%%%%%%%%%%%%%%%
\begin{figure}
    \centering
    \includegraphics[width=0.92\linewidth, height=5cm]{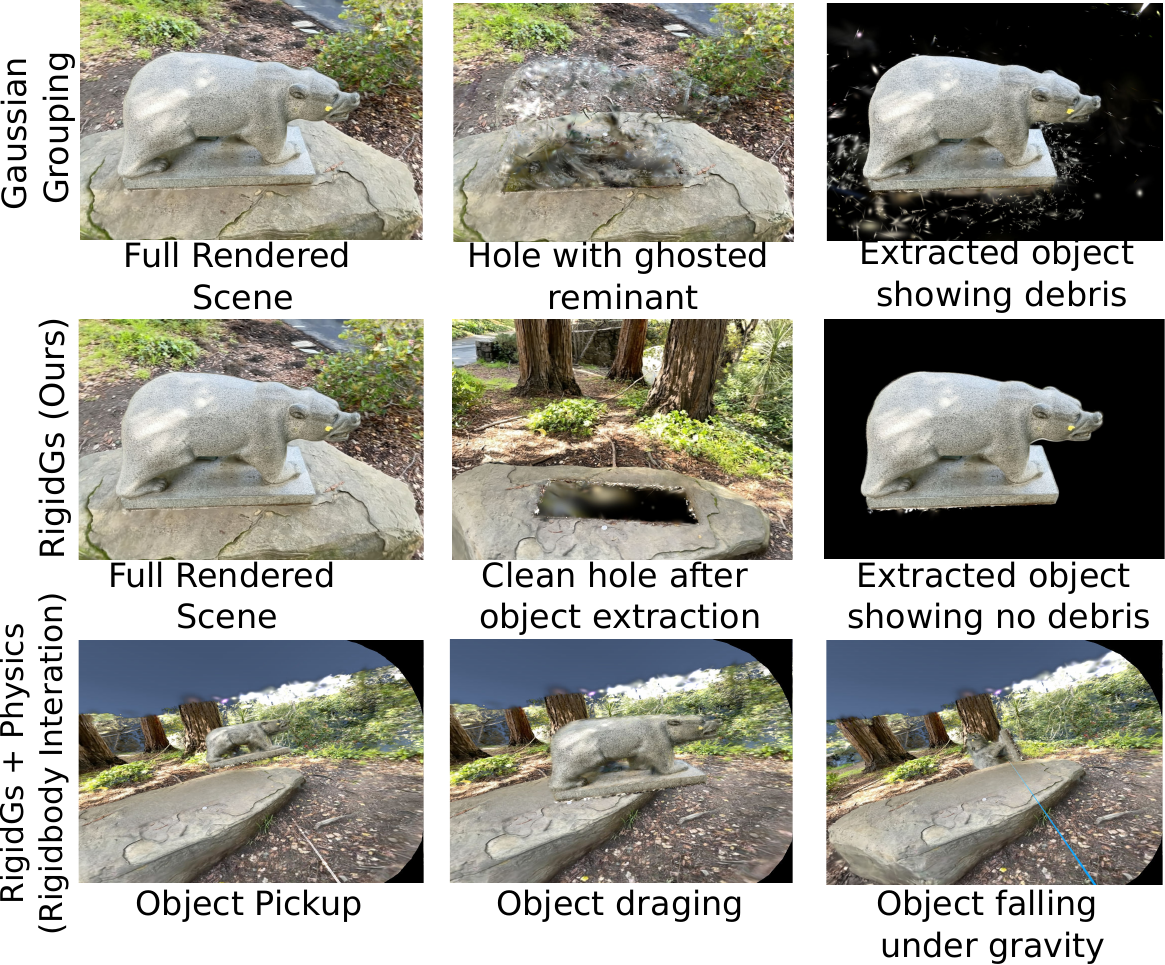}
    \caption{\rev{\textbf{Object extraction comparison on \textit{Bear scene from IN2N dataset}~\cite{mip-nerf-360}.} RigidGS produces clean object separation with minimal to zero remnant debris or artifacts, enabling direct, mesh-free rigid-body manipulation of Gaussian primitives.}}
    \vspace{-10pt}
    \label{fig:overview}
\end{figure}
%%%%%%%%%%%%%%%%%%%%%%%%%%%%%%%%%%%%%%%%%%%%%%%%%%%%%%%%%%%%%%%%%%%%%%%%%%%%%%%%%%%%%%%%%%%%%%%%%%
\section{Introduction}
\label{sec:intro}

Virtual and Mixed Reality (VR/MR) applications require photorealistic 3D scene representations that support real-time rendering and object-level interaction~\cite{vr-gs, vr-splat, gs-verse, xin2025a3fr}. While 3D Gaussian Splatting (3DGS)~\cite{3dgs} achieves high-fidelity novel-view synthesis with efficient rasterization, it models scenes as a monolithic set of primitives lacking persistent object membership, preventing the object-level manipulation, removal, and physics-based interaction that immersive environments require~\cite{vr-gs, wang2025decoupledgaussian}.

Recent approaches add object structure by learning semantic state over the Gaussian field, via identity embeddings~\cite{gaussian-grouping}, linear programming~\cite{flashsplat}, graph cuts~\cite{gaussiancut}, lifted foundation features~\cite{cen2025segment}, boundary-adaptive attributes~\cite{Zhang_2025_CVPR}, or anchor collections~\cite{zhu2025objectgs}. These methods rest on two assumptions that often lead to poor segmentation: that 2D foundation models provide accurate mask supervision, and that semantic identity can be safely co-optimized with geometry. SAM2 frequently errs on complex geometry, showing \emph{over-inclusion} on thin, elongated structures and \emph{under-inclusion} on multimodal-depth objects~\cite{Zhang_quantifying}. Moreover, co-optimizing identity with geometry induces a \emph{label--geometry coupling}: semantic gradients depend on the geometry of non-converged neighbors, pulling boundary primitives toward erroneous object ownership.

We present \textbf{RigidGS}, an end-to-end framework that decouples geometric learning from semantic identity. RigidGS first corrects the two SAM2 failure modes with a depth- and temporally-aware refinement pipeline: neighboring refined masks are flow-warped into each view and, on disagreement with raw SAM2, trigger re-prompting from depth-consistent donor candidates, after which a strict-grid KNN depth threshold (via Depth-Anything-V2) and a ridge-fill step recover thin structures and remove depth-inconsistent regions. Each Gaussian is then labeled once by COLMAP-projection voting on the refined masks~\cite{colmap} with a spatial KNN cleanup, and these labels are \emph{frozen} before geometric optimization: a 2D mask BCE loss supervises only geometry and appearance while identities propagate unchanged through densification and pruning. Determining identity entirely upstream eliminates the label--geometry coupling that affects methods with learnable per-Gaussian semantic state.

Running COB-GS~\cite{Zhang_2025_CVPR} with our refined masks substantially closes its background-bleeding gap on multimodal-depth scenes ($+9.5$ IoU on \textit{trex}; \cref{tab:maskimpact}), yet a residual gap remains because identity is co-optimized: learnable-identity methods avoid collapse only through reactive mechanisms such as COB-GS's boundary-adaptive splitting, at the cost of extra primitives and residual bleed, whereas our fixed-identity design removes the coupling by construction. On LLFF/NVOS~\cite{ren2022neural} RigidGS improves IoU by $\mathbf{7.7}$ and boundary IoU by $\mathbf{14.5}$ points over COB-GS, and also leads on Mip-NeRF~360~\cite{mip-nerf-360} and iNeRF360. RigidGS runs at interactive rates on a Quest~2 deployment. We complement the trained representation with LaMa-based background inpainting and refinement~\cite{lama, spin-nerf, wang2026inpaint360gs, lee2026gpgs}, and demonstrate rigid-body interaction via affine transforms on Gaussian subsets, using game-engine physics rather than baking expensive physics-based models~\cite{vr-gs, physgaussian}.

\medskip\noindent\textbf{Contributions.}
\begin{enumerate}[leftmargin=*,itemsep=1pt,topsep=2pt]
\item A depth- and temporally-aware mask refinement preprocessor that
resolves SAM2 \emph{over-} and \emph{under-inclusion}, as a drop-in module
for 2D-supervised 3DGS methods.
\item A decoupled 3DGS formulation in which object identity is fixed upstream
and frozen, restricting the mask loss to geometric optimization and
eliminating label--geometry coupling by construction.
\item An end-to-end 3DGS VR system supporting real-time rigid-body
manipulation and inpainting, deployed on a Meta Quest~2.
\end{enumerate}
%%%%%%%%%%%%%%%%%%%%%%%%%%%%%%%%%%%%%%%%%%%%%%%%%%%%%%%%%%%%%%%%%%%%%%%%%%%%%%%%%%%%%%%%%%%%%%%%%%

\begin{figure*}[!ht]
\centering
\includegraphics[width=0.99\linewidth]{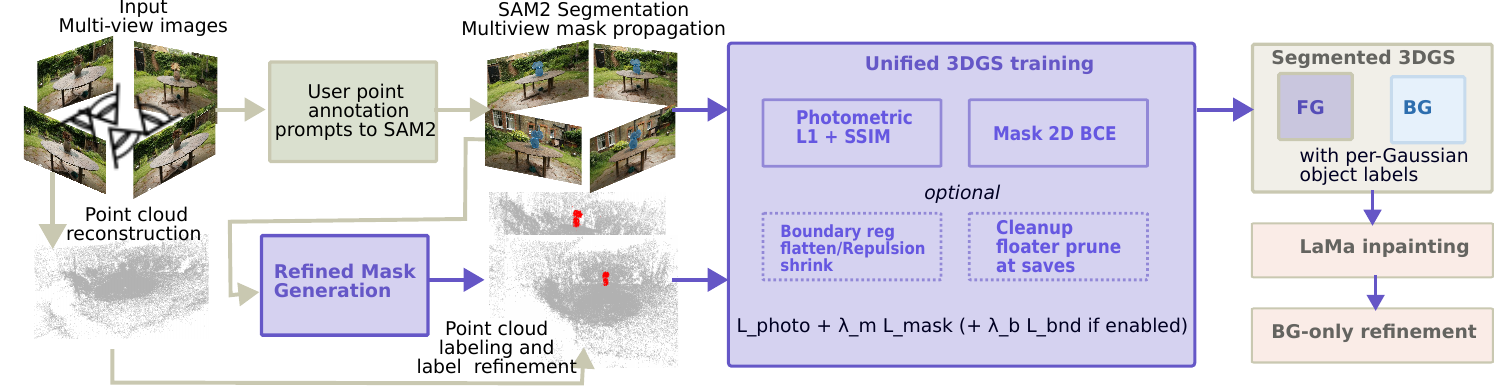}

\caption{Overview of the RigidGS pipeline.}
\vspace{-10pt}
\label{fig:RigidGS Pipeline}
\end{figure*}
\section{Related Work}

\noindent\textbf{3D Gaussian Splatting for immersive rendering.}
NeRF~\cite{nerf} introduced coordinate-based volumetric scene representations, but its ray-marching formulation makes real-time interaction difficult. Later work improved scalability and speed through anti-aliasing, unbounded scenes, hash encodings, and structured grids~\cite{mip-nerf,mip-nerf-360,instant-ngp,yu_and_fridovichkeil2021plenoxels,sun2022direct,chen2022tensorf,k-planes}. 3D Gaussian Splatting (3DGS)~\cite{3dgs} shifted this paradigm toward rasterized anisotropic primitives, enabling high-quality real-time rendering. Subsequent methods improve surface quality, stability, hierarchy, compression, and streaming~\cite{2dgs,sugar,scafoldgs,minigs,radl2024stopthepop,huang2024error,niedermayr2024compressed,fan2024lightgaussian,instantstream}. VR-oriented variants further target foveation, low-latency reprojection, and headset-efficient splatting~\cite{fan2025fov,vr-splatting,xin2025a3fr,vr-splat}. These works make Gaussian scenes increasingly suitable for immersive display, but they do not directly address object ownership and manipulation.

\noindent\textbf{Object-aware Gaussian scenes.}
Recent methods extend 3DGS with semantic structure by attaching learnable identity features, embeddings, or mask attributes to Gaussians~\cite{gaussian-grouping,cen2025segment,feature-3dgs, Zhang_2025_CVPR,zhu2025objectgs,zhang2025labelgs}, formulating assignment via optimization~\cite{flashsplat,gaussiancut}, or enabling interactive/language-guided editing~\cite{wu2024gaussctrl,gaussianeditor,3dsceneeditor,xu2024texture,wang2024view,featuresplatting,point-and-move,shi2026realm,zhao2025isegman}. However, jointly optimizing identity with geometry under noisy 2D masks induces conflicting boundary supervision, causing background primitives near edges or interior gaps to drift into the foreground. RigidGS eliminates this coupling by refining masks upstream, freezing object labels before training, and using downstream losses strictly for geometric optimization.

\noindent\textbf{Mask refinement and object removal.}
Object-aware 3DGS pipelines typically inherit supervision from 2D foundation models (\eg, Grounded-SAM, SAM, SAM2)~\cite{ren2024grounded,liu2024grounding,sam,sam2}, which remain brittle on thin, low-contrast, or complex structures~\cite{Zhang_quantifying}. While 2D refiners (\eg, HQ-SAM, SAMRefiner)~\cite{saiq,lin2025samrefiner} and temporal aggregators (\eg, DEVA, SAM2)~\cite{cheng2023tracking,sam2} improve mask quality, 3D-aware methods leverage multi-view cues~\cite{sanerfhq,wang2026sgs}. RigidGS operates in this latter category as a lightweight preprocessor: depth-quantile re-prompting and strict-grid refinement resolve both over- and under-inclusion \emph{before} 3D optimization. Freezing these masks upstream ensures clean primitive separation, critical for downstream background inpainting and removal~\cite{spin-nerf,lama,wang2026inpaint360gs,lee2026gpgs}.

\noindent\textbf{Physical interaction with Gaussian scenes.}
Several works connect Gaussian representations to dynamic simulation and interactive environments, including physics-aware deformation, fluid and elastic effects, mesh- or cage-based control, and MLLM-guided simulation~\cite{physgaussian,vr-gs,gaussiansplashing,pig,reconstruction,meshbased,gs-verse,gsdeformer,gasp,physsplat,wang2025decoupledgaussian,wu20244d,duan20244d}. These approaches often require specialized solvers, deformation models, or material assumptions. RigidGS instead targets the common VR case of rigid object manipulation: once object Gaussians are cleanly separated, grasping, translating, and rotating objects can be implemented as affine transformations over Gaussian subsets and integrated directly with game-engine physics.

%%%%%%%%%%%%%%%%%%%%%%%%%%%%%%%%%%%%%%%%%%%%%%%%%%%%%%%%%%%%%%%%%%%%%%%%%%%%%%%%%%%%%%%%%%%%%%%%%%

\section{Method}
\label{sec:method}

\subsection{Preliminaries}
\label{sec:prelim}

We briefly review 3D Gaussian Splatting (3DGS)~\cite{3dgs}. A scene is represented as a set of anisotropic Gaussian primitives $\mathcal{G} = {g_i}_{i=1}^{N}$. Each is parameterized by a 3D mean $\boldsymbol{\mu}_i \in \mathbb{R}^3$, scale $\mathbf{s}_i \in \mathbb{R}^3$, rotation quaternion $\mathbf{q}_i$, opacity $\alpha_i \in [0,1]$, and spherical harmonic coefficients encoding view-dependent color. The covariance is $\boldsymbol{\Sigma}_i = \mathbf{R}_i \mathbf{S}_i \mathbf{S}_i^{\top} \mathbf{R}_i^{\top}$, where $\mathbf{R}_i$ comes from $\mathbf{q}_i$ and $\mathbf{S}_i = \mathrm{diag}(\mathbf{s}_i)$. Gaussians are rasterized via front-to-back alpha blending. The entire rendering is differentiable in $\boldsymbol{\mu}_i, \mathbf{s}_i, \mathbf{q}_i, \alpha_i$ under a photometric objective combining $\ell_1$ and SSIM losses. RigidGS retains this rendering pipeline and adds three components. These include a SAM2 mask-refinement step (\cref{sec:maskrefine}), a fixed-identity formulation that determines object membership entirely upstream of geometric optimization (\cref{sec:identity}), and a two-stage schedule with background inpainting (\cref{sec:background}). \Cref{fig:RigidGS Pipeline} gives the overview.

\subsection{Depth- and Temporally-Aware Mask Refinement}
\label{sec:maskrefine}

Raw SAM2~\cite{sam2} masks degrade via two failure modes. First, \emph{over-inclusion} occurs when gaps in thin, elongated structures (\eg, ferns, branches) fall below the model's resolution, thereby absorbing background into the foreground mask. Second, \emph{under-inclusion} occurs when objects spanning multimodal depth bands (\eg, a truck cab and bed) are truncated to the primary depth band. Because RigidGS fixes object identities at initialization (\cref{sec:identity}), preprocessing mask quality is necessary for accurate point labeling.

To resolve both modes, RigidGS generates refined masks using SAM2 output masks $\{M_v^{\text{SAM2},k}\}$ frame by frame.  Sparse user clicks on the first frame are propagated by SAM2 to all views, producing raw masks ${M_v^{\text{SAM2},k}}$. For each subsequent frame, recently refined masks from neighboring views are warped into the current view via dense optical flow~\cite{teed2020raft}; if the raw SAM2 mask disagrees with the warped consensus, re-prompting is triggered. Re-prompt points are sampled from the mask interior and candidate regions within a depth quantile band $[q_\alpha, q_{1-\alpha}]$ from Depth-Anything-V2~\cite{depthanything}. The quantile filter preserves multimodal depth structures rather than rejecting secondary bands as outliers (resolving under-inclusion). Candidates are validated via local template matching. Finally, a spatial grid applies a KNN soft depth threshold under local color/connectivity constraints to eliminate background bleed (addressing over-inclusion), followed by ridge-filling to output the refined mask $M_v^{\text{ref},k}$.

\subsection{Fixed Object Identity and Mask Supervision of Geometry}
\label{sec:identity}
\label{sec:maskloss}

RigidGS treats object identity as a fixed attribute determined entirely before geometric optimization. Given refined masks ${M_v^{\text{ref},k}}$, a sparse COLMAP point cloud~\cite{colmap} is constructed. For each 3D point $\mathbf{x}$ visible in views $\mathcal{V}(\mathbf{x})$, the label is assigned by majority vote across views:
\begin{equation}
y(\mathbf{x}) = \arg\max_{c \in {0,\dots,K}} \sum_{v \in \mathcal{V}(\mathbf{x})} \mathbbm{1}\!\left[\pi_v(\mathbf{x}) \in M_v^{\text{ref},c}\right],
\end{equation}
Here, $c=0$ denotes background. A one-shot spatial KNN voting pass then reassigns isolated stragglers to the majority label of their local 3D neighborhood. After cleanup, each Gaussian carries a constant integer label $y_i \in {0,\dots,K}$. This partitions $\mathcal{G} = \bigcup_{k=0}^{K} \mathcal{G}_k$ with $\mathcal{G}_k = {g_i \mid y_i = k}$. Labels are never exposed to any gradient. Cloned and split Gaussians inherit $y_i$, pruned Gaussians are removed together with their labels, and no training-time loss can modify them.

\textbf{Why avoid in-training identity refinement?}
Methods like COB-GS~\cite{Zhang_2025_CVPR}, ObjectGS~\cite{zhu2025objectgs}, and LabelGS~\cite{zhang2025labelgs} optimize per-Gaussian semantic states via gradient descent, exposing object identities to gradients computed from partially converged neighbor geometry. This coupling could be harmful even over short durations: as little as 10 iterations of logit refinement can introduce foreground primitives into the background. Upstream refined-mask voting provides sufficient label accuracy, making in-training identity updates unneeded and eliminating label--geometry coupling entirely.

\textbf{Mask supervision of geometry.}
Although frozen identities cleanly partition Gaussians, residual 2D mask noise and 3D primitive overlap can cause boundary leakage. We resolve this using a projection-consistent 2D mask loss that forces boundary primitives to align with refined masks across views. For each view $v$, rendering foreground Gaussians ($y_i \neq 0$) as white and background ($y_i = 0$) as black produces a soft rasterized occupancy map $\hat{M}_v$. We optimize geometry via Binary Cross-Entropy against refined masks:
\begin{equation}
\mathcal{L}_{\text{mask}} = \frac{1}{|\mathcal{V}|}\sum_{v \in \mathcal{V}} \mathrm{BCE}(\hat{M}_v, M_v^{\text{ref}}).
\end{equation}
Because identities $y_i$ are frozen constants, gradients flow exclusively into geometric parameters ($\boldsymbol{\mu}_i, \mathbf{s}_i, \mathbf{q}_i, \alpha_i$). The loss acts strictly as a geometric regularizer: misplaced primitives are adjusted in position, scale, or opacity without altering identity.

To prevent shape erosion from minor silhouette under-recovery, we apply a distance transform to mask background pixels within a scene-scaled distance $\rho_b$ of the boundary, setting $\mathcal{L}_{\text{mask}} = 0$ in this transition band while fully supervising pixels beyond it. Margins ($\rho_b$) are scaled narrowest for thin, intricate structures (\eg, fern fronds).

\subsection{Boundary Stabilization}
\label{sec:boundary}

The mask BCE provides image-space supervision but does not directly prevent interpenetration between Gaussians of different objects in 3D. We add a lightweight boundary stabilizer that operates on the cross-class interface. Boundary Gaussians are identified from a $k$-NN graph on detached positions. A Gaussian $i$ is in the boundary set $\mathcal{B}$ if at least $\tau$ of its neighbors carry a different label, and cross-class neighbor pairs form a deduplicated set $\mathcal{P}$. For each $i \in \mathcal{B}$, an oriented flattening term penalizes variance along the estimated interface direction $\mathbf{n}i$. This encourages thin discs aligned with the interface. A scale-shrinkage hinge penalizes any boundary Gaussian whose maximum scale exceeds $s{\max}^{\star}$. A cross-class repulsion hinge with margin $m$ separates pair centers:
$$
\begin{aligned}
\mathcal{L}_{\text{boundary}}
={}&\lambda_f\frac{1}{|\mathcal{B}|}\sum_{i\in\mathcal{B}}
\mathbf{n}_i^\top\boldsymbol{\Sigma}_i\mathbf{n}_i
+\lambda_s\frac{1}{|\mathcal{B}|}\sum_{i\in\mathcal{B}}
\max\!\left(0,\max(\mathbf{s}_i)-s_{\max}^{\star}\right)\\
&+\lambda_r\frac{1}{|\mathcal{P}|}\sum_{(i,j)\in\mathcal{P}}
\max\!\left(0,m-\|\boldsymbol{\mu}_i-\boldsymbol{\mu}_j\|_2\right).
\end{aligned}
$$

Boundary stabilization is scene-dependent: it gives large gains on thin or elongated structures (e.g., $+0.286$ IoU on \textit{fern}) but has little effect on compact silhouettes such as \textit{fortress} or \textit{horns}. This is expected, since it acts locally at cross-class interfaces and therefore scales with boundary complexity.

\begin{figure*}[t]
    \centering
    \includegraphics[width=0.9\linewidth, height=6.5cm]{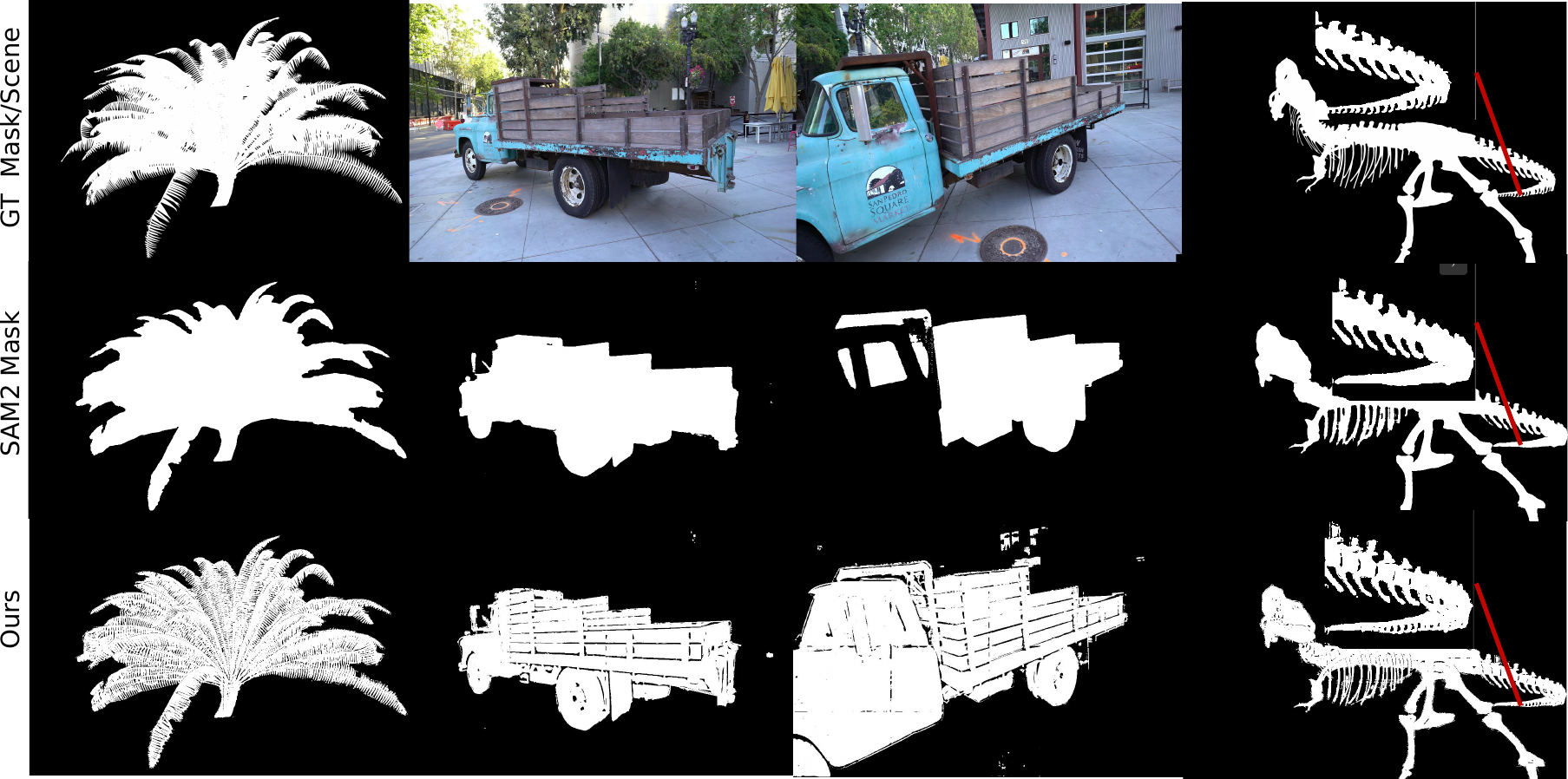}
    \caption{Mask refinement. Top: raw SAM2; bottom: ours. \textit{Fern}
(over-inclusion), \textit{truck} (under-inclusion), \textit{trex} (both).}
    \vspace{-10pt}
    \label{fig:mask_refinement}
\end{figure*}

\subsection{Training and Background Inpainting}
\label{sec:background}

Object removal exposes occluded regions that lack direct photometric supervision: the original images require the object to be present, whereas completing the background requires targets in which it is absent. We resolve this conflict via a two-stage schedule. In Stage~1 ($T_1$ iterations), the scene is optimized using photometric, mask BCE, and boundary-stabilization losses, with frozen object identities from \cref{sec:identity}. At iteration $T_{\text{inpaint}}$, we mask out foreground Gaussians; occluded regions are propagated across views via SAM2 and inpainted with LaMa~\cite{lama} to yield background targets $\{I_v^{\text{inpaint}}\}$. As few background Gaussians lie behind the removed object, we first seed $\mathcal{G}_0$ inside the hole at depths interpolated from the surrounding background. In Stage~2, optimization alternates between full-scene photometric loss on original images and a background-only loss:
\begin{equation}
\mathcal{L}_{\text{bg}} = \frac{1}{|\mathcal{V}|}\sum_{v \in \mathcal{V}} \bigl\| \hat{I}_v^{\text{bg}} - I_v^{\text{inpaint}} \bigr\|_1,
\end{equation}
with gradients restricted to $\mathcal{G}_0$, which is further densified near the hole (details in the supplementary). Since $\mathcal{L}_{\text{bg}}$ spans the whole frame, regions outside the hole remain anchored to the Stage~1 background, and completion leaves foreground geometry untouched.

\subsection{Rigid-Body Manipulation}
\label{sec:rigid}

Each fixed-identity subset $\mathcal{G}_k$ is moved as a rigid body by an affine transformation $(\mathbf{R}, \mathbf{t})$ applied to its constituent Gaussians: $\boldsymbol{\mu}_i’ = \mathbf{R}\boldsymbol{\mu}_i + \mathbf{t}$, $\mathbf{R}_i’ = \mathbf{R}\mathbf{R}_i$. The relative geometry and covariance structure are preserved, and the fixed identities ensure that each subset moves coherently and independently. No mesh extraction, deformation field, or retraining is required; objects remain native Gaussian groups throughout manipulation and integrate directly with standard game-engine physics.
%%%%%%%%%%%%%%%%%%%%%%%%%%%%%%%%%%%%%%%%%%%%%%%%%%%%%%%%%%%%%%%%%%%%%%%%%%%%%%%%%%%%%%%%%%%%%%%%%%

\section{Experiments}
\label{sec:experiments}

We evaluate RigidGS along six axes: the quality of refined masks
(\cref{sec:exp-masks}), object segmentation against prior work
(\cref{sec:exp-segmentation}), the impact of refined masks on 3DGS
segmentation methods (\cref{sec:exp-maskimpact}), component analysis
through ablation (\cref{sec:exp-ablation}), object removal and
background completion (\cref{sec:exp-removal}), and rigid-body
interaction (\cref{sec:exp-rigid}).

\subsection{Datasets, Implementation, and Protocol}
\label{sec:datasets}

\textbf{Datasets.}
We evaluate RigidGS across diverse 3DGS benchmarks. The set includes
LLFF/NVOS~\cite{ren2022neural} scenes (\textit{fern}, \textit{flower},
\textit{fortress}, \textit{horns}, \textit{orchids}, \textit{trex}),
Mip-NeRF 360~\cite{mip-nerf-360} scenes (\textit{garden},
\textit{kitchen}), the Instruct-NeRF2NeRF~\cite{nerf2nerf}
\textit{bear} scene, and Tanks and Temples (T and T)~\cite{tank_and_temples}
scenes (\textit{truck}, \textit{train}). Together these cover thin
structures, cluttered indoor scenes, complex silhouettes, and
multimodal-depth foregrounds.

\textbf{Implementation.}
RigidGS is built on the standard 3DGS pipeline~\cite{3dgs} with adaptive
density control. We train for 30{,}000 iterations on the original views,
trigger background inpainting at iteration 30{,}000, and continue with
7{,}000 background-refinement iterations. SAM2~\cite{sam2} is prompted
once per object with sparse user clicks on the first frame; masks are
propagated through the multi-view set and refined by our pipeline
before COLMAP-projection voting. All baselines are run with their
published code.

\textbf{Metrics.}
We report mask intersection-over-union (IoU), boundary IoU (bIoU) restricted to a thin band around the true boundary, and pixel accuracy.

\textbf{Evaluation protocol.} All methods are evaluated on the same held-out test views, ground-truth images, and ground-truth masks. Each ground-truth view is paired with the rendered view closest to it in photometric content. We pair by content rather than file name because, for some LLFF scenes, the NVOS evaluation masks are annotated on frames whose names do not directly match our held-out test images; the pairing does not depend on the reported segmentation metrics. For SPIn-NeRF removal, we compute all metrics only on the inpainted region against the object-free ground truth: outside the provided object mask, the render is replaced by the ground truth, PSNR/SSIM are averaged over mask pixels, and LPIPS (AlexNet) is computed on the mask's bounding-box crop. We re-score Gaussian Grouping under this identical protocol.

\subsection{Mask Refinement Quality}
\label{sec:exp-masks}

\Cref{fig:mask_refinement} shows raw SAM2 masks and our refined masks
on three representative scenes. \textit{Fern} (over-inclusion): SAM2
absorbs background gaps between fronds into the foreground silhouette;
our strict-grid local depth refinement recovers them without sacrificing
thin-structure continuity. \textit{Truck} (under-inclusion): SAM2
concentrates on the cab and excludes the differently-depthed bed; our
donor-warp re-prompting with quantile depth candidates recovers the
bed. \textit{Trex} (combined): the silhouette has thin ribs requiring
over-inclusion correction and a tail at a different depth band requiring
under-inclusion recovery.

\subsection{Object Segmentation Against Prior Work}
\label{sec:exp-segmentation}
We compare RigidGS against Gaussian Grouping~\cite{gaussian-grouping},
SAGA~\cite{cen2025segment}, and COB-GS~\cite{Zhang_2025_CVPR}. Gaussian
Grouping is restricted to \textit{fern}, \textit{fortress}, and
\textit{horns} of LLFF because it fragments thin or compound objects on
other scenes.

\begin{table}[t]
\centering
\footnotesize
\caption{Object segmentation quality averaged over scenes per dataset. Per-scene results appear
in the supplementary material.}
\label{tab:main_segmentation}
\setlength{\tabcolsep}{3pt}
\renewcommand{\arraystretch}{0.75}
\begin{tabular}{@{}llccc@{}}
\toprule
Dataset & Method & IoU $\uparrow$ & bIoU $\uparrow$ & Acc $\uparrow$ \\
\midrule
\multirow{4}{*}{LLFF / NVOS}
& Gaussian Grouping & 0.631 & 0.269 & 0.866 \\
& SAGA & 0.665 & 0.357 & 0.898 \\
& COB-GS & 0.742 & 0.576 & 0.951 \\
& \textbf{RigidGS} & \textbf{0.819} & \textbf{0.721} & \textbf{0.961} \\
\midrule
\multirow{4}{*}{Mip-NeRF 360}
& Gaussian Grouping & 0.266 & 0.180 & 0.802\\
& SAGA & 0.892 & 0.691 & 0.987 \\
& COB-GS & 0.927 & 0.777 & 0.993 \\
& \textbf{RigidGS} & \textbf{0.951} & \textbf{0.846} & \textbf{0.994} \\
\midrule
\multirow{4}{*}{IN2N (Bear)}
& Gaussian Grouping & 0.552 & 0.221 & 0.795 \\
& COB-GS & 0.956 & 0.760 & 0.989 \\
& SAGA & 0.971 & 0.826 & 0.992 \\
& \textbf{RigidGS} & \textbf{0.980} & \textbf{0.877} & \textbf{0.995} \\
\midrule
\multirow{4}{*}{Tanks \& Temples}
& Gaussian Grouping & 0.485 & 0.223 & 0.722 \\
& SAGA & 0.820 & 0.346 & 0.935 \\
& RigidGS & 0.945 & 0.725 & 0.983 \\
& \textbf{COB-GS} & \textbf{0.968} & \textbf{0.779} & \textbf{0.989} \\
\bottomrule
\end{tabular}
\vspace{-5pt}
\end{table}

\Cref{tab:main_segmentation} reports aggregate metrics per dataset.
On LLFF/NVOS, RigidGS outperforms COB-GS by 7.7 IoU (0.819 vs.\ 0.742) and 14.5 bIoU (0.721 vs.\ 0.576), and leads on Mip-NeRF 360 and \textit{IN2N} (0.980 IoU), confirming our fixed-identity formulation suppresses fragmentation on thin, complex structures. While COB-GS slightly leads IoU on Tanks \& Temples (0.968 vs.\ 0.945) via boundary-adaptive splitting, standard IoU ignores foreground residue attached to the background (see qualitative results in supplementary material). RigidGS achieves cleaner large-object removal with minimal foreground residue, ensuring the clean isolation required for downstream rigid-body manipulation.

\begin{figure*}[t]
    \centering
    \includegraphics[width=0.89\linewidth]{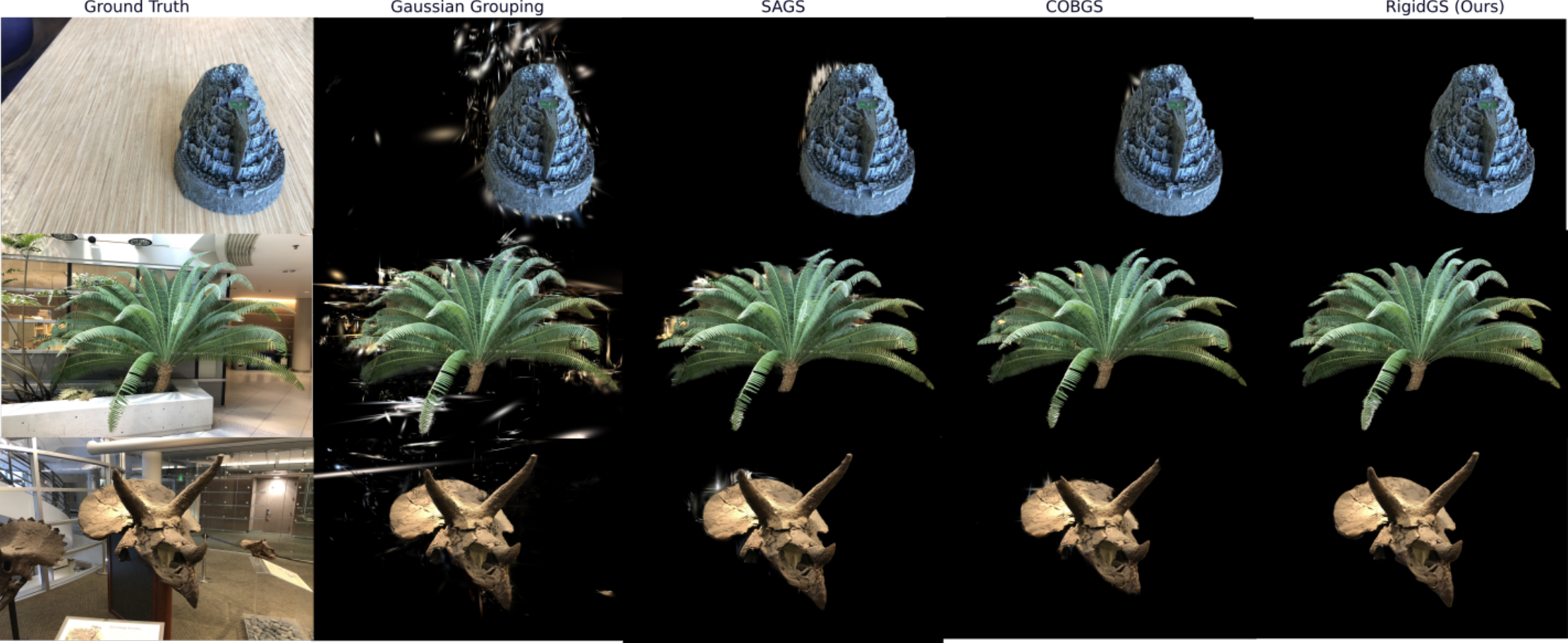}
    \caption{Object segmentation comparison across LLFF, Mip-NeRF 360,
    iNeRF360, and Tanks \& Temples scenes.}
    \label{fig:separation_qual}
    \vspace{-10pt}
\end{figure*}

Qualitative comparisons are shown in
\cref{fig:separation_qual}: Gaussian Grouping and SAGA produce
fragmented extractions on thin structures, COB-GS retains background
drift on multimodal-depth objects, and RigidGS produces clean
segmentation across scene types.

\begin{table}[t]
\centering
\small
\caption{Multi-object segmentation on LERF scenes.}
\label{tab:multiobject}
\begin{tabular}{lccccc}
\toprule
 & & \multicolumn{2}{c}{Ours} & \multicolumn{2}{c}{GG} \\
\cmidrule(lr){3-4} \cmidrule(lr){5-6}
Scene & \#Obj & IoU & bIoU & IoU & bIoU \\
\midrule
Figurines & 10 & 0.96 & 0.82 & 0.37 & 0.20 \\
Teatime   & 5  & 0.97 & 0.97 & 0.22 & 0.13 \\
\bottomrule
\end{tabular}
\end{table}

    \textbf{Multi-Object Segmentation.} We evaluate RigidGS on two multi-object LERF scenes~\cite{kerr2023lerf}, figurines (10 objects) and teatime (5 objects), achieving 0.96 / 0.97 IoU versus 0.37 / 0.22 for Gaussian Grouping (Table~\ref{tab:multiobject}); qualitative comparisons are provided in the supplementary material.

\subsection{Impact of Mask Refinement on 3DGS Segmentation}
\label{sec:exp-maskimpact}
To isolate our mask refinement from the fixed-identity formulation, we evaluate a $2 \times 2$ grid across all six LLFF/NVOS scenes: {COB-GS, RigidGS} $\times$ {native masks, our refined masks}. COB-GS uses Grounded-SAM2~\cite{liu2024grounding, ren2024grounded} (text-prompted DINO + SAM2) as its native pipeline, and point-prompted SAM2~\cite{sam2} with sparse clicks for RigidGS.

\begin{table}[t]
\centering
\footnotesize
\caption{Effect of mask refinement on 3DGS segmentation across all
LLFF/NVOS scenes.}

\label{tab:maskimpact}
\setlength{\tabcolsep}{3pt}
\renewcommand{\arraystretch}{0.65}
\begin{tabular}{@{}llccc@{}}
\toprule
Scene & Configuration & IoU $\uparrow$ & bIoU $\uparrow$ & Acc $\uparrow$ \\
\midrule
\multirow{4}{*}{Fern}
& COB-GS (Grounded-SAM2) & 0.747 & 0.688 & \textbf{0.916} \\
& COB-GS + refined & 0.733 & 0.694 & 0.912 \\
& RigidGS (SAM2) & \textbf{0.753} & 0.323 & 0.915 \\
& RigidGS + refined & 0.726 & \textbf{0.714} & 0.907 \\
\midrule
\multirow{4}{*}{Flower}
& COB-GS (Grounded-SAM2) & 0.952 & 0.833 & 0.989 \\
& COB-GS + refined & 0.954 & 0.831 & 0.989 \\
& RigidGS (SAM2) & 0.963 & \textbf{0.854} & 0.991 \\
& RigidGS + refined & \textbf{0.968} & 0.792 & \textbf{0.993} \\
\midrule
\multirow{4}{*}{Fortress}
& COB-GS (Grounded-SAM2) & 0.967 & 0.656 & 0.994 \\
& COB-GS + refined & 0.964 & 0.612 & 0.993 \\
& RigidGS (SAM2) & 0.978 & 0.869 & 0.996 \\
& RigidGS + refined & \textbf{0.985} & \textbf{0.905} & \textbf{0.997} \\
\midrule
\multirow{4}{*}{Horns}
& COB-GS (Grounded-SAM2) & 0.876 & 0.589 & 0.978 \\
& COB-GS + refined & 0.970 & 0.809 & 0.995 \\
& RigidGS (SAM2) & 0.975 & 0.859 & \textbf{0.996} \\
& RigidGS + refined & \textbf{0.978} & \textbf{0.877} & \textbf{0.996} \\
\midrule
\multirow{4}{*}{Orchids}
& COB-GS (Grounded-SAM2) & 0.414 & 0.243 & 0.898 \\
& COB-GS + refined & 0.427 & 0.260 & 0.872 \\
& RigidGS (SAM2) & 0.661 & 0.468 & \textbf{0.935} \\
& RigidGS + refined & \textbf{0.666} & \textbf{0.485} & \textbf{0.935} \\
\midrule
\multirow{4}{*}{Trex}
& COB-GS (Grounded-SAM2) & 0.496 & 0.446 & 0.931 \\
& COB-GS + refined & 0.591 & 0.556 & \textbf{0.939} \\
& RigidGS (SAM2) & \textbf{0.598} & \textbf{0.559} & \textbf{0.939} \\
& RigidGS + refined & 0.590 & 0.550 & 0.938 \\
\bottomrule
\end{tabular}
\end{table}

\Cref{tab:maskimpact} reports quantitative results across all LLFF scenes; a qualitative $2{\times}2$ comparison is provided in the supplementary material).

Key findings to note: First, mask refinement primarily boosts COB-GS on thin or multimodal-depth scenes suffering from under-inclusion (\textit{horns}: $+9.4$ IoU, $+22.0$ bIoU; \textit{trex}: $+9.5$ IoU; LLFF average: $+3.1$ IoU, $+5.1$ bIoU), proving mask quality bottlenecks joint optimization. Second, RigidGS’s IoU is largely mask-invariant ($0.821$ vs.\ $0.819$ refined) because projection voting and KNN cleanup enforce 3D consistency. Instead, refinement targets boundary precision, lifting LLFF average bIoU from $0.655$ to $0.721$, most visibly on \textit{fern} ($0.323 \to 0.714$) by removing over-included inter-frond regions. Third, refinement addresses under-inclusion for COB-GS (mass) and over-inclusion for RigidGS (boundaries), proving it is complementary to our fixed-identity formulation rather than redundant.

While refinement yields the largest gains on severe base-mask failures such as under-inclusion on multi-depth objects (\textit{horns}, \textit{trex} for COB-GS) or over-inclusion between thin structures (\textit{fern} for RigidGS), on clean base masks (\textit{flower}), it offers minimal benefit and can over-tighten boundaries (RigidGS bIoU $0.854 \to 0.792$). Close, unrelated structures can also corrupt the mask (building metals close to \textit{trex} corrupting the mask, making RigidGS IoU $0.598 \to 0.590$). Refinement is thus best reserved for thin, complex, or multi-depth geometry.

\subsection{Component Analysis}
\label{sec:exp-ablation}

We evaluate four RigidGS variants on LLFF scenes: (i) \textit{Init only} (COLMAP voting + KNN cleanup, no 2D loss); (ii) \textit{$+$ BCE} (2D mask loss without boundary stabilization); (iii) \textit{Full RigidGS} (our complete pipeline); and (iv) \textit{Full $+$ learnable logits} (unfrozen per-Gaussian logits). \Cref{tab:ablation_iou} reports segmentation IoU; bIoU follows identical trends (see supplementary).

\begin{table}[t]
\centering
\caption{\textbf{Component analysis on LLFF scenes (IoU).} Boundary stabilization and frozen logits are critical for complex, multi-depth geometry (\textit{trex}, \textit{fern}).}
\label{tab:ablation_iou}
\small
\setlength{\tabcolsep}{4pt}
\renewcommand{\arraystretch}{0.95}
\resizebox{\linewidth}{!}{%
\begin{tabular}{@{}lccccc@{}}
\toprule
Variant & Fern & Fortress & Horns & Trex & Avg \\
\midrule
Init only (COLMAP+KNN)            & 0.485 & 0.676 & 0.820 & 0.324 & 0.576 \\
$+$ BCE (no boundary)            & 0.444 & 0.962 & \textbf{0.978} & 0.287 & 0.668 \\
\textbf{Full RigidGS (ours)}     & 0.726 & \textbf{0.985} & \textbf{0.978} & \textbf{0.590} & \textbf{0.820} \\
Full $+$ learnable logits        & \textbf{0.730} & 0.962 & 0.975 & 0.466 & 0.783\\
\bottomrule
\end{tabular}%
}
\vspace{-5pt}
\end{table}

\textbf{Unstabilized 2D BCE destabilizes thin/multimodal scenes.} Adding 2D mask BCE without boundary stabilization drops IoU below the initialization baseline on \textit{fern} ($0.485 \to 0.444$) and \textit{trex} ($0.324 \to 0.287$). Without 3D regularization, 2D silhouette agreement forces boundary Gaussians to drift inconsistently across views. On simple silhouettes (\textit{fortress}, \textit{horns}), BCE alone suffices ($0.676 \to 0.962$, $0.820 \to 0.978$).

\textbf{Boundary stabilization is essential for complex geometry.} Stabilization recovers performance on \textit{fern} ($0.444 \to 0.726$) and \textit{trex} ($0.287 \to 0.590$) while remaining neutral on compact silhouettes. Its impact scales with the surface-area-to-volume ratio of the target object.

\textbf{Label–geometry coupling collapses multi-depth scenes.} Re-enabling learnable logits matches our frozen formulation on unimodal-depth scenes, but causes a 21\% relative drop on \textit{trex} ($0.590 \to 0.466$ IoU; $0.550 \to 0.439$ bIoU). Because \textit{trex} features strongly multimodal foreground depth (ribs vs.\ tail), multiple geometric convergence modes induce identity gradients that pull boundary Gaussians toward inconsistent labels. Freezing logits avoids this optimization trap.

\subsection{Robustness to initial label noise.} To evaluate sensitivity to frozen initialization errors, we corrupt COLMAP-projection labels by flipping 10–30\% of primitives (bg$\leftrightarrow$fg and symmetric). Final IoU is virtually invariant up to 30\% corruption on \textit{fern}, \textit{trex}, and \textit{kitchen} ($<0.004$ deviation; \cref{tab:noise_robustness}), while \textit{garden} shows a small non-progressive offset ($0.963 \to 0.928$). Tracing mislabeled Gaussians on \textit{kitchen} scene revealed that 2D mask loss suppresses their opacity by 5$\times$ within one densification cycle (median $0.092$ vs.\ $0.486$) and an order of magnitude by iteration 1000. Standard opacity pruning then physically removes 97.9\% of mislabeled primitives ($39{,}685 \to 836$). Rather than relabeling, RigidGS purges corrupt Gaussians entirely. Downstream object removal shows slightly higher noise sensitivity (see the supplementary material).

\begin{table}[t]
\centering
\small
\caption{Robustness to initial label corruption.}
\label{tab:noise_robustness}
\setlength{\tabcolsep}{4pt}
\renewcommand{\arraystretch}{0.75}
\begin{tabular}{@{}lcccc@{}}
\toprule
Corruption & Fern & Trex & Garden & Kitchen \\
\midrule
None (baseline) & 0.726 & 0.590 & 0.963 & 0.939 \\
\midrule
bg$\to$fg 10\% & 0.727 & 0.591 & 0.928 & 0.942 \\
bg$\to$fg 20\% & 0.726 & 0.592 & 0.928 & 0.941 \\
bg$\to$fg 30\% & 0.727 & 0.591 & 0.928 & 0.941 \\
\midrule
fg$\to$bg 10\% & 0.727 & 0.593 & 0.928 & 0.942 \\
fg$\to$bg 20\% & 0.729 & 0.593 & 0.928 & 0.942 \\
fg$\to$bg 30\% & 0.730 & 0.593 & 0.928 & 0.942 \\
\midrule
symmetric 10\% & 0.728 & 0.593 & 0.928 & 0.942 \\
symmetric 20\% & 0.729 & 0.591 & 0.928 & 0.942 \\
symmetric 30\% & 0.730 & 0.592 & 0.927 & 0.941 \\
\bottomrule
\end{tabular}
\vspace{-5pt}
\end{table}

\subsection{Object Removal and Background Completion}
\label{sec:exp-removal}

\begin{figure}[h]
    \centering
    \includegraphics[width=0.99\linewidth]{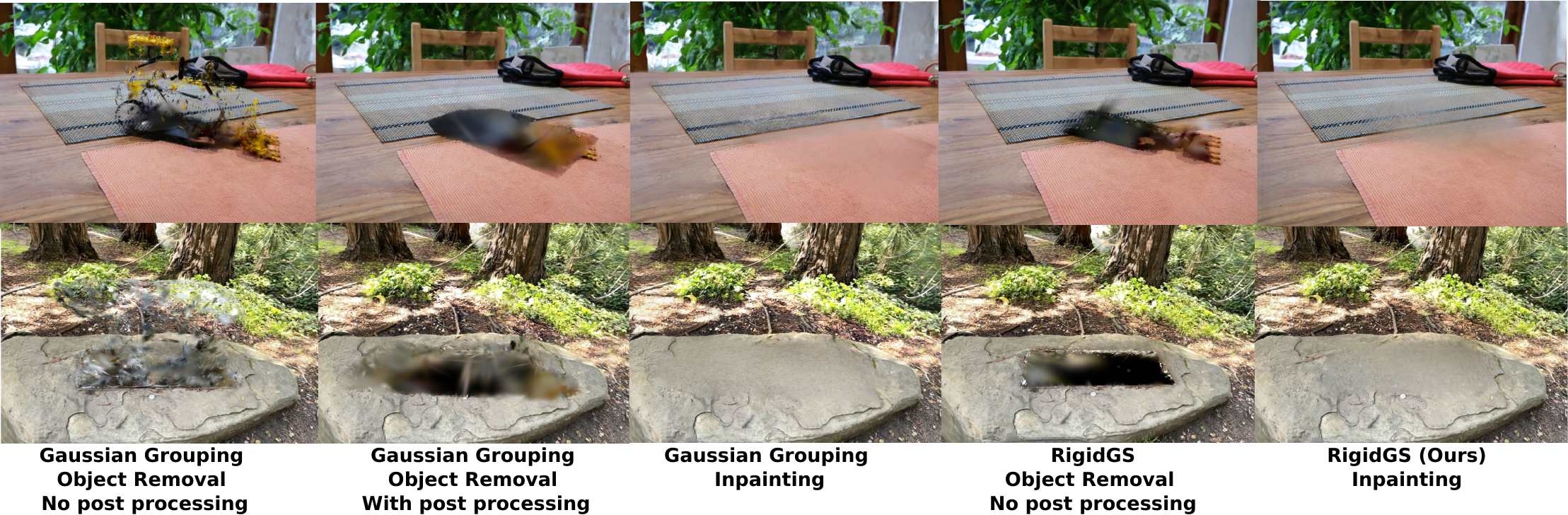}
    \caption{Object removal and background reconstruction. RigidGS
    produces clean removal with view-consistent background recovery
    while Gaussian Grouping retains ghost residue at the vacated
    region.}
    \label{fig:inpainting}
    \vspace{-10pt}
\end{figure}

\Cref{fig:inpainting} illustrates object removal. Gaussian Grouping leaves ghosting residue in vacated regions because low-opacity foreground primitives persist in its learnable identity field. By integrating precomputed inpainting directly into 3DGS optimization (\cref{sec:background}), RigidGS achieves view-consistent recovery without post-hoc cleanup. Crucially, a mislabeled Gaussian visibly degrades object extraction, whereas an erroneously removed background primitive leaves a hole that our inpainting stage can easily recover.

We evaluate background completion inside the removal mask on three SPIn-NeRF~\cite{spin-nerf} scenes with object-free ground truth (\cref{tab:removal}). RigidGS outperforms Gaussian Grouping by \textbf{+2.4\,dB PSNR} and \textbf{+0.235 SSIM}, restoring more faithful background structure. However, Gaussian Grouping outperforms RigidGS on LPIPS and produces smoothed content that favors perceptual metrics.

\begin{table}[t]
\centering

\footnotesize
\setlength{\tabcolsep}{6pt}
\renewcommand{\arraystretch}{0.95}

\caption{\textbf{SPIn-NeRF object removal}~\cite{spin-nerf}. In-mask metrics
vs.\ object-free ground truth; RigidGS leads on PSNR and SSIM.}
\label{tab:removal}
\begin{tabular}{@{}lccc@{}}
\toprule
Method & PSNR $\uparrow$ & SSIM $\uparrow$ & LPIPS $\downarrow$ \\
\midrule
Gaussian Grouping & 14.04 & 0.217 & \textbf{0.583} \\
\textbf{RigidGS} & \textbf{16.44} & \textbf{0.452} & 0.602 \\
\bottomrule
\end{tabular}
\vspace{-10pt}
\end{table}

\subsection{Rigid-Body Interaction}
\label{sec:exp-rigid}

Each segmented object is exported as a standalone Gaussian subset $\mathcal{G}_k$ with a bound physics collider. Rigid motion is executed by applying real-time affine transforms over constituent primitives, enabling gravity, collision response, dragging, and placement without mesh extraction, deformation fields, or retraining. Because our fixed-identity design produces clean object boundaries free of foreground debris, primitives are passed directly to the physics engine without post-hoc cleanup. We report quantitative rendering and manipulation latency in \cref{tab:deployment}; please refer to the supplementary material and video for qualitative interaction sequences (\eg, free fall, inter-object collisions, user dragging) across the \textit{garden} and \textit{bear} scenes.

\subsection{Deployment and Computational Cost}
\label{sec:exp-deployment}

We further evaluate RigidGS across two deployment axes: real-time rendering/interaction on a VR headset (\cref{tab:deployment}) and offline scene construction cost (\cref{tab:cost}).

\textbf{Real-time VR deployment.} On a Meta Quest~2, RigidGS maintains $\sim$33\,FPS during static rendering and $\sim$26\,FPS during active manipulation on complex scenes (\eg, \textit{garden} with 4.03M Gaussians; \textit{bear} with 2.93M Gaussians). Because standalone 3DGS typically requires sub-million-primitive counts to sustain native VR refresh rates, frame rates are bottlenecked by mobile GPU rasterization rather than by our framework. High-framerate alternatives require specialized rendering paths (\eg, foveation, compact primitives~\cite{vr-splatting, vr-splat}) or desktop tethering~\cite{vr-splat}. Operating within standard 3DGS rasterization constraints, RigidGS introduces per-subset affine transforms, adding only 0.19\,ms/frame on average (0.25\, ms at p95). RigidGS thus enables object-level rigid-body manipulation as an additive capability on top of monolithic 3DGS with negligible overhead.

\textbf{Computational cost.} \Cref{tab:cost} breaks down scene construction overhead. Our core contribution, depth- and temporally-aware mask refinement, adds just 1--3\, min of automated processing, with sparse user prompt clicks taking $\sim$1\, min/object. Offline costs are dominated by standard 3DGS training ($\sim$2--3\,h on an RTX A6000) and background inpainting ($\sim$10--13\,min), shared with prior object-aware methods. The training overhead relative to boundary-free baselines stems from per-iteration nearest-neighbor queries in the boundary regularizer; this cost is strictly offline and does not affect runtime. Peak GPU memory remains low (9.5\, GB) because object identities are stored as frozen integer labels per Gaussian, introducing zero extra parameters, gradients, or optimizer states.

\begin{table}[t]
\centering
\small
\setlength{\tabcolsep}{4pt}
\renewcommand{\arraystretch}{0.9}
\caption{\textbf{Meta Quest~2 deployment.} FPS is a
90-frame rolling average; transform latency is the per-subset affine
update overhead.}
\vspace{-5pt}
\label{tab:deployment}
\begin{tabular}{@{}lcc@{}}
\toprule
Metric & Garden & Bear \\
\midrule
Gaussian count          & 4.03M    & 2.93M \\
Static FPS (mean)        & 32.7     & 32.8 \\
Interaction FPS (mean)   & 25.6     & 26.6 \\
Transform latency (avg)  & 0.19\,ms & 0.19\,ms \\
Transform latency (p95)  & 0.23\,ms & 0.25\,ms \\
\bottomrule
\end{tabular}
\vspace{-5pt}
\end{table}

\begin{table}[t]
\centering
\footnotesize
\setlength{\tabcolsep}{4pt}
\renewcommand{\arraystretch}{0.9}
\caption{\textbf{RigidGS computational cost} (RTX A6000)}

\label{tab:cost}
\begin{tabular}{@{}lcc@{}}
\toprule
Stage & Fern & Trex \\
\midrule
\multicolumn{3}{@{}l}{\textit{Preprocessing (one-time)}} \\
\quad Annotation (user clicks)   & 62\,s    & 86\,s \\
\quad Segmentation (automated)   & 15\,s    & 69\,s \\
\quad Mask refinement (ours)     & 1.2\,min & 3.1\,min \\
\midrule
\multicolumn{3}{@{}l}{\textit{Training (30k iters)}} \\
\quad Wall-clock                 & 2.6\,h   & 2.2\,h \\
\quad Peak GPU memory            & 9.5\,GB  & 9.5\,GB \\
\midrule
\multicolumn{3}{@{}l}{\textit{Background completion (one-time)}} \\
\quad Inpainting + finetuning    & 10.0\,min & 13.4\,min \\
\bottomrule
\end{tabular}
\vspace{-15pt}
\end{table}
%%%%%%%%%%%%%%%%%%%%%%%%%%%%%%%%%%%%%%%%%%%%%%%%%%%%%%%%%%%%%%%%%%%%%%%%%%%%%%%%%%%%%%%%%%%%%%%%%%
\section{Conclusion}
\textbf{Limitations.} 
First, offline training takes 2--3\, hours per scene because the boundary regularizer requires nearest-neighbor graph updates; future work should explore caching or approximate queries to speed this up. Second, rendering on a Quest~2 reaches $\sim$26--33\, FPS (below the 72\, Hz VR target) due to millions of Gaussian counts; however, because the manipulation overhead is sub-millisecond, future work should explore implementing RigidGS using specialized VR renderers that employ foveation/primitive pruning.

RigidGS enables real-time, object-level manipulation in 3DGS by decoupling semantic identity from geometric optimization. By combining depth- and temporally-aware mask refinement with frozen projection voting, RigidGS eliminates the label--geometry coupling inherent in co-optimization frameworks. Across LLFF/NVOS, Mip-NeRF~360, and iNeRF360, RigidGS significantly improves object segmentation and boundary fidelity over prior baselines. The resulting clean primitive partitions enable mesh-free rigid-body interaction at interactive frame rates on a Meta Quest~2. Future work will optimize offline boundary training costs and extend the framework to articulated and non-rigid dynamics.

%%%%%%%%%%%%%%%%%%%%%%%%%%%%%%%%%%%%%%%%%%%%%%%%%%%%%%%%%%%%%%%%%%%%%%%%%%%%%%%%%%%%%%%%%%%%%%%%%%
\section{Acknowledgment}
This work was supported in part by the AI-Seed Research Award from the Research and Economic Development Division at the University of Wyoming. %In addition, research supported by the National Science Foundation (NSF) under Grant No. 2340997 contributed to the development of ideas in this work.
{
    \small
    \bibliographystyle{ieeenat_fullname}
    \bibliography{wacv_main}
}

\end{document}